\documentclass[runningheads]{llncs}
\makeatletter
\def\input@path{{styles/}}
\makeatother

\usepackage[T1]{fontenc}
\usepackage[utf8]{inputenc}
\usepackage{graphicx}
\usepackage{hyperref}
\usepackage{amsmath}
\usepackage{cite}
\usepackage{pgfplots}
\pgfplotsset{compat=1.18}
\usepackage{booktabs}
\usepackage{xspace}

\newcommand{\hrefurl}[1]{\href{#1}{\fontsize{10}{11}\selectfont#1}}

\title{RDF-Based Structured Quality Assessment Representation of Multilingual LLM Evaluations}
\titlerunning{RDF-Based Quality Assessment of Multilingual LLMs}

\author{Jonas Gwozdz \and Andreas Both}
\authorrunning{J. Gwozdz and A. Both}
\institute{Leipzig University of Applied Sciences, Leipzig, Germany\\
\email{\{jonas.gwozdz,andreas.both\}@htwk-leipzig.de}}

\begin{document}

\maketitle

\begin{abstract}
Large Language Models (LLMs) increasingly serve as knowledge interfaces, yet systematically assessing their reliability with conflicting information remains difficult. 
We propose an RDF-based framework to assess multilingual LLM quality, focusing on knowledge conflicts. 
Our approach captures model responses across four distinct context conditions (complete, incomplete, conflicting, and no-context information) in German and English. 
This structured representation enables the comprehensive analysis of knowledge leakage-where models favor training data over provided context-error detection, and multilingual consistency. 
We demonstrate the framework through a fire safety domain experiment, revealing critical patterns in context prioritization and language-specific performance, and demonstrating that our vocabulary was sufficient to express every assessment facet encountered in the 28-question study.
\end{abstract}

\section{Introduction}
\begin{figure}[t]
\centering
\includegraphics[width=1\linewidth]{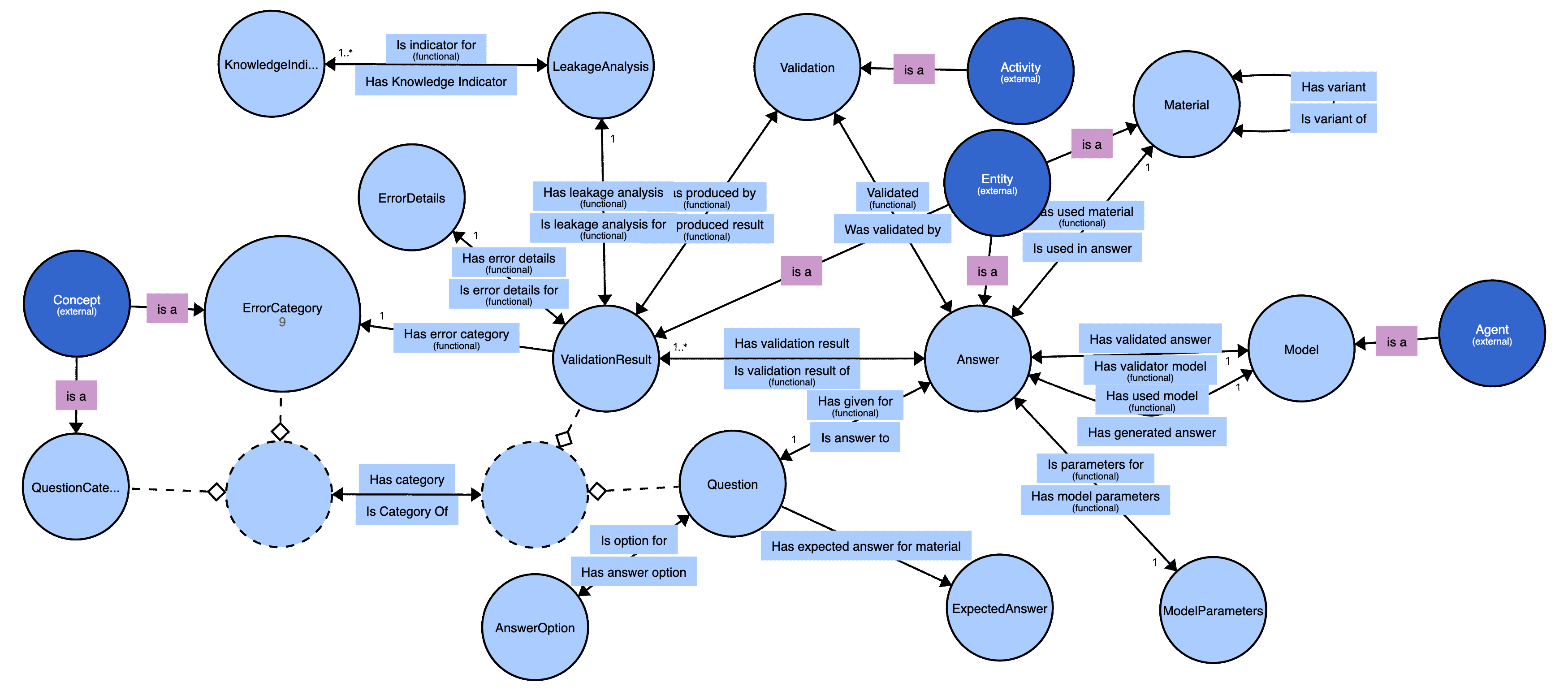}
\vspace*{-3ex}
\caption{Simplified visual representation of the RDF vocabulary for LLM evaluations.}
\label{fig:schema}
\vspace*{-3ex}
\end{figure}

As interfaces to vast amounts of knowledge, Large Language Models (LLMs) are increasingly prevalent in information processing. However, their tendency to blend knowledge from training data with provided context poses challenges for reliability assessment, particularly in critical domains where factual accuracy is essential \cite{lavrinovics2024knowledgegraphslargelanguage}. With incomplete or conflicting information, LLMs may prioritize either training data or provided context, impacting their reliability.
Current evaluation approaches often lack standardized, structured representations of assessment results, hindering systematic analysis and comparison. This gap is especially pronounced in multilingual evaluations, where performance can differ across languages, and in understanding the interplay between context-based and training-based knowledge, which remains underexplored. Moreover, without adhering to FAIR principles (Findable, Accessible, Interoperable, Reusable), evaluation results are poorly materialized, limiting their usability and broader adoption in research and practice.
We address these challenges by introducing an RDF-based framework for the structured representation of LLM quality assessments across models and languages. Our approach offers several key contributions:
(1) a comprehensive RDF vocabulary for representing LLM evaluation results that aligns with FAIR principles to ensure findability, accessibility, interoperability, and reusability,
(2) a systematic methodology for testing LLM responses across four distinct context conditions—complete, incomplete, conflicting, and no-context information—supporting multilingual evaluations with language-specific extensions to capture model- and language-specific behaviors,
and (3) a demonstration validating our data model through a fire safety domain experiment with models like GPT-4o-mini\footnote{\hrefurl{https://openai.com/index/gpt-4o-mini-advancing-cost-efficient-intelligence/}} and Gemini-2.0-Flash\footnote{\hrefurl{https://developers.googleblog.com/en/gemini-2-family-expands/}} in German and English, exposing critical patterns in context prioritization and language-specific performance within the resulting dataset. 
Figure~\ref{fig:schema} illustrates the core structure of our RDF vocabulary, which underpins the evaluation framework.

\section{Related Work}
Recent studies explore LLMs with knowledge graphs and knowledge conflicts. 
Lavrinovics et al.~\cite{lavrinovics2024knowledgegraphslargelanguage} survey how knowledge graphs mitigate LLM hallucinations, while Kwan et al.~\cite{kwan2024using} use them for factual accuracy. 
Xie et al.~\cite{xie2024adaptivechameleonstubbornsloth} and Tan et al.~\cite{tan2024blindedgeneratedcontextslanguage} highlight LLMs' bias toward context, even when incorrect. 
However, these efforts lack standardized vocabularies for systematic, multilingual evaluation.
While valuable, current approaches often focus on narrow contexts or fail to provide structured, interoperable frameworks aligned with FAIR principles. Our work addresses this limitation by introducing an RDF-based representation for multilingual LLM assessments, enabling consistent, queryable analysis of knowledge conflicts and language-specific behaviors across varied context scenarios.

\newcommand{\rdf}[1]{\texttt{:#1}}

\section{RDF Vocabulary for LLM Evaluation}
Our RDF vocabulary links \rdf{Question}, \rdf{Answer}, \rdf{ValidationResult}, and\linebreak[4]{} \rdf{Material} to evaluate LLMs across languages and context conditions (complete, incomplete, conflicting, no-context). 
Multilinguality is supported via language-tagged literals for \rdf{hasText}. 
Relationships such as \rdf{hasGivenFor} (\rdf{Question} to \rdf{Answer}), \rdf{hasUsedMaterial} (\rdf{Answer} to \rdf{Material}), and\linebreak[3]{} \rdf{hasValidation-}\linebreak[3]{}\texttt{Result} (\rdf{Answer} to \rdf{ValidationResult}) enable SPARQL queries for analyzing knowledge leakage and cross-lingual consistency. 
Figure~\ref{fig:schema} illustrates this core structure with key relationships.

The T-Box defines 14 classes (e.g., \rdf{Question}, \rdf{Answer}) and 57 properties (e.g., \rdf{isValid}, \rdf{matchesFactual}), with OWL/SHACL constraints ensuring integrity.
It aligns with FAIR principles using the PROV Ontology\footnote{\hrefurl{https://www.w3.org/TR/prov-o/}} and Dublin Core\footnote{\hrefurl{https://www.dublincore.org/specifications/dublin-core/dcmi-terms/}}.
The full schema is available at \url{http://purl.org/sqare#}.
The choice of RDF enables reasoning, linkage to external KGs, and federated SPARQL queries---advantages unattainable with flat CSV tables.

\section{Use Case and Evaluation}
We applied our approach to a fire safety domain experiment, showcasing its ability to structure LLM quality assessments. 
Fire safety was chosen due to its well-defined knowledge base and critical need for accuracy, making it ideal for testing the schema's capacity to capture context variations as well as multilingual responses.
Beyond this domain, the framework generalizes to applications like educational content creation.
By testing LLM responses against course materials, creators can pinpoint knowledge gaps and ensure content sufficiency, with the RDF structure tracking material-question alignments. 
Full experimental details are available in the online appendix (Git repository).

\subsubsection{Experimental Setup}
The experiment tested LLM responses to 28 fire safety questions under four context conditions: (1) \emph{Complete}: full, accurate information; (2) \emph{Incomplete}: missing information; (3) \emph{Conflicting}: factually contradicting information; (4) \emph{No Context}: no supporting context. These conditions were chosen to assess how LLMs prioritize context versus training knowledge, revealing behaviors like context adherence or knowledge leakage.
All prompts follow a \emph{zero-shot, system-first} template detailed in the online appendix.

\paragraph{Data Collection and Analysis}
For each question and context condition, we collected responses from GPT-4o-mini and Gemini-2.0-Flash in both German and English, storing them in our RDF structure. 
Validation assessed correctness per fire safety standards and context expectations. 
The resulting RDF dataset captures anomalies and notable LLM behaviors, such as deviations from expected results, comprehensively reflecting the assessed LLMs' capabilities.
SPARQL queries enabled the analysis of knowledge patterns, multilingual differences, and context reliance across models and languages.

\paragraph{Key Findings}
Tables~\ref{tab:stats_german} and~\ref{tab:stats_english} present the performance of GPT-4o-mini and Gemini-2.0-Flash in German and English under four context conditions:
%
%
%
\emph{Context Dominance}: All models strongly adhere to the given instructions and prioritize provided context, replicating incorrect information at rates of 89-93\% rather than using their training knowledge;
\emph{Multilingual Differences}: English models handle incomplete information better, while German models demonstrate stronger baseline knowledge without context, revealing language-specific behaviors; 
\emph{Educational Applications}: For course creators, these findings suggest focusing on information completeness in English materials while potentially leveraging German LLMs' stronger baseline knowledge for gap identification.

All such findings are reflected in the vocabulary.
Hence, such findings can be generated using SPARQL queries\footnote{See our online appendix at \url{http://purl.org/sqare/repo\#}.}, which validates our research task of representing the data from LLM assessments in a comprehensive and semantically rich form.

\subsection{Paired Statistical Comparison of Models}
To complement our RDF-based evaluation, we performed a paired analysis on the binary correctness labels (\texttt{is\_valid}) for each question under each context (\emph{complete}, \emph{incomplete}, \emph{conflicting}, \emph{no\_context}) and language (\emph{de}, \emph{en}).  For each model pair (Gemini-2.0-Flash vs.\ GPT-4o-mini) we built the 2×2 contingency table
\[
\begin{bmatrix}
a & b\\
c & d
\end{bmatrix}
=
\begin{bmatrix}
\text{both correct} & \text{Gemini correct, GPT wrong}\\
\text{Gemini wrong, GPT correct} & \text{both wrong}
\end{bmatrix}
\]
and computed:
\begin{enumerate}
  \item \textbf{McNemar's exact test} (two-sided, no continuity correction) on the discordant cells \(b\) vs.\ \(c\).
  \item \(\Delta\)\textbf{–accuracy} \(=\) Acc\(_{\rm Gemini}\)\(-\)Acc\(_{\rm GPT}\), with 95 \% Newcombe CI for paired proportions.
  \item \textbf{Cohen's \(\kappa\)} as a measure of overall agreement beyond chance.
\end{enumerate}

\begin{table}[t]
\centering
\vspace*{-2ex}
\caption{Paired Statistical Comparison (German): Gemini-2.0-Flash vs.\ GPT-4o-mini}
\label{tab:stats_german}
\scriptsize
\begin{tabular}{lcccc}
\toprule
\textbf{Context} & \textbf{Contingency (a,b;c,d)} & \textbf{McNemar \(p^\dagger\)} & \textbf{\(\Delta\)-Acc (95 \% CI) [pp]} & \textbf{Cohen's \(\kappa\)} \\
\midrule
Complete     & (28, 0; 0, 0) & -          & \(0.0\) [0.0,\,0.0]            & - ($\kappa$ undefined)      \\
Incomplete   & (10, 4; 8, 6) & 0.3877          & \(-14.3\) [\(-37.9\),\,+9.4] & 0.143  \\
Conflicting  & ( 2, 0; 1,25) & -          & \(-3.6\) [\(-10.4\),\,+3.3] & 0.781  \\
No Context   & (24, 2; 2, 0) & -          & \(0.0\) [\(-14.0\),\,+14.0] & $-0.077$ \\
\bottomrule
\end{tabular}

\raggedright\scriptsize
$^\dagger$ Exact two-sided McNemar $p$-value; “–” indicates b + c < 5
\vspace*{-2ex}
\end{table}
\begin{table}[t!]
\centering
\vspace*{-2ex}
\caption{Paired Statistical Comparison (English): Gemini-2.0-Flash vs.\ GPT-4o-mini}
\label{tab:stats_english}
\scriptsize
\begin{tabular}{lcccc}
\toprule
\textbf{Context} & \textbf{Contingency (a,b;c,d)} & \textbf{McNemar \(p^\dagger\)} & \textbf{\(\Delta\)-Acc (95 \% CI) [pp]} & \textbf{Cohen's \(\kappa\)} \\
\midrule
Complete     & (28, 0; 0, 0) & -          & \(0.0\) [0.0,\,0.0]            & - ($\kappa$ undefined)      \\
Incomplete   & (27, 1; 0, 0) & -          & \(+3.6\) [\(-3.3\),\,+10.4] & 0      \\
Conflicting  & ( 2, 1; 1,24) & -          & \(0.0\) [\(-9.9\),\,+9.9]  & 0.627  \\
No Context   & (14, 0; 9, 5) & 0.0039 & \(-32.1\) [\(-49.4\),\,-14.8] & 0.357  \\
\bottomrule
\end{tabular}

\raggedright\scriptsize
$^\dagger$ Exact two-sided McNemar $p$-value; “–” indicates b + c < 5
\vspace*{-2ex}
\end{table}

McNemar's test is non-significant (\(p > 0.05\)) in all German contexts, including \emph{incomplete} (\(p = 0.3877\)). In other cases such as \emph{no\_context}, the number of discordant pairs was too low (\(b + c < 5\)) to permit meaningful testing, despite a noticeable accuracy gap of \(-14.3\) percentage points in \emph{incomplete}.
In English, only the \emph{no\_context} condition shows a significant discordance (\(p=0.0039\)), with GPT-4o-mini outperforming Gemini by 32.1 pp.  The Newcombe CIs reveal that many of these gaps are too wide to draw firm conclusions (e.g., German \emph{incomplete} CI spans ±25 pp), while Cohen's \(\kappa\) highlights very high agreement in error replication (\(\kappa=0.781\) de, 0.627 en) versus low agreement under manipulated prompts (e.g., \(\kappa=0.143\) de, undefined in en).

\section{Conclusion and Future Work}

Our RDF vocabulary (contribution 1) and fire safety experiment (contribution 2) assess LLM reliability across languages and contexts, focusing on knowledge conflicts critical for real-world use.
Experiments with GPT-4o-mini and Gemini-2.0-Flash in German and English reveal how models handle contradictory information and language-specific differences, emphasizing the need for structured knowledge.
Findings show that LLMs favor training data over context in 7-11\% of conflicting cases, but mostly adhere to the given context, even when incorrect.

This semantically rich representation ensures reliability in critical systems, supporting standardized, queryable analysis for transparency and reproducibility.
It advances structured LLM evaluation methodologies, tackling reliability challenges with conflicting information across languages.
Future work will scale the question set, add evaluations for low-resource languages, and refine leakage metrics, as well as extend this framework to new domains, improve knowledge leakage detection, and define reliability metrics.

\bibliographystyle{splncs04}
\bibliography{references}

\end{document}